\documentclass[final]{cvpr}

\usepackage{times}
\usepackage{epsfig}
\usepackage{graphicx}
\usepackage{amsmath}
\usepackage{amssymb}
\usepackage{caption,subcaption}
\usepackage{comment}
\usepackage{amssymb}
\usepackage{pifont}
\newcommand{\cmark}{\ding{51}}%
\newcommand{\xmark}{\ding{55}}%
\usepackage{enumerate}
\usepackage{enumitem}
\usepackage[pagebackref=true,breaklinks=true,colorlinks,bookmarks=false]{hyperref}

\def\cvprPaperID{6055} 
\def\confYear{CVPR 2021}

\begin{document}

\title{Robust Instance Segmentation through\\ Reasoning about Multi-Object Occlusion}

\author{Xiaoding Yuan$^{1}$ \qquad Adam Kortylewski$^{2}$\qquad Yihong Sun$^{2}$ \qquad Alan Yuille$^{2}$\\
\\
$^1$Tongji University, $^2$Johns Hopkins University \\
}

\maketitle
\pagestyle{empty}  
\thispagestyle{empty} 

\begin{abstract}

Analyzing complex scenes with Deep Neural Networks is a challenging task, particularly when images contain multiple objects that partially occlude each other. 
Existing approaches to image analysis mostly process objects independently and do not take into account the relative occlusion of nearby objects.
In this paper, we propose a deep network for multi-object instance segmentation that is robust to occlusion and can be trained from bounding box supervision only.
Our work builds on Compositional Networks, which learn a generative model of neural feature activations to locate occluders and to classify objects based on their non-occluded parts. 
We extend their generative model to include multiple objects and introduce a framework for efficient inference in challenging occlusion scenarios.
In particular, we obtain feed-forward predictions of the object classes and their instance and occluder segmentations.
We introduce an Occlusion Reasoning Module (ORM) that locates erroneous segmentations and estimates the occlusion order to correct them.
The improved segmentation masks are, in turn, integrated into the network in a top-down manner to improve the image classification.
Our experiments on the KITTI INStance dataset (KINS) and a synthetic occlusion dataset
demonstrate the effectiveness and robustness of our model at multi-object instance segmentation under occlusion. 
Code is publically available at \href{https://github.com/XD7479/Multi-Object-Occlusion}{https://github.com/XD7479/Multi-Object-Occlusion}.
\end{abstract}
\begin{figure}
    \centering
    \includegraphics[width=\linewidth]{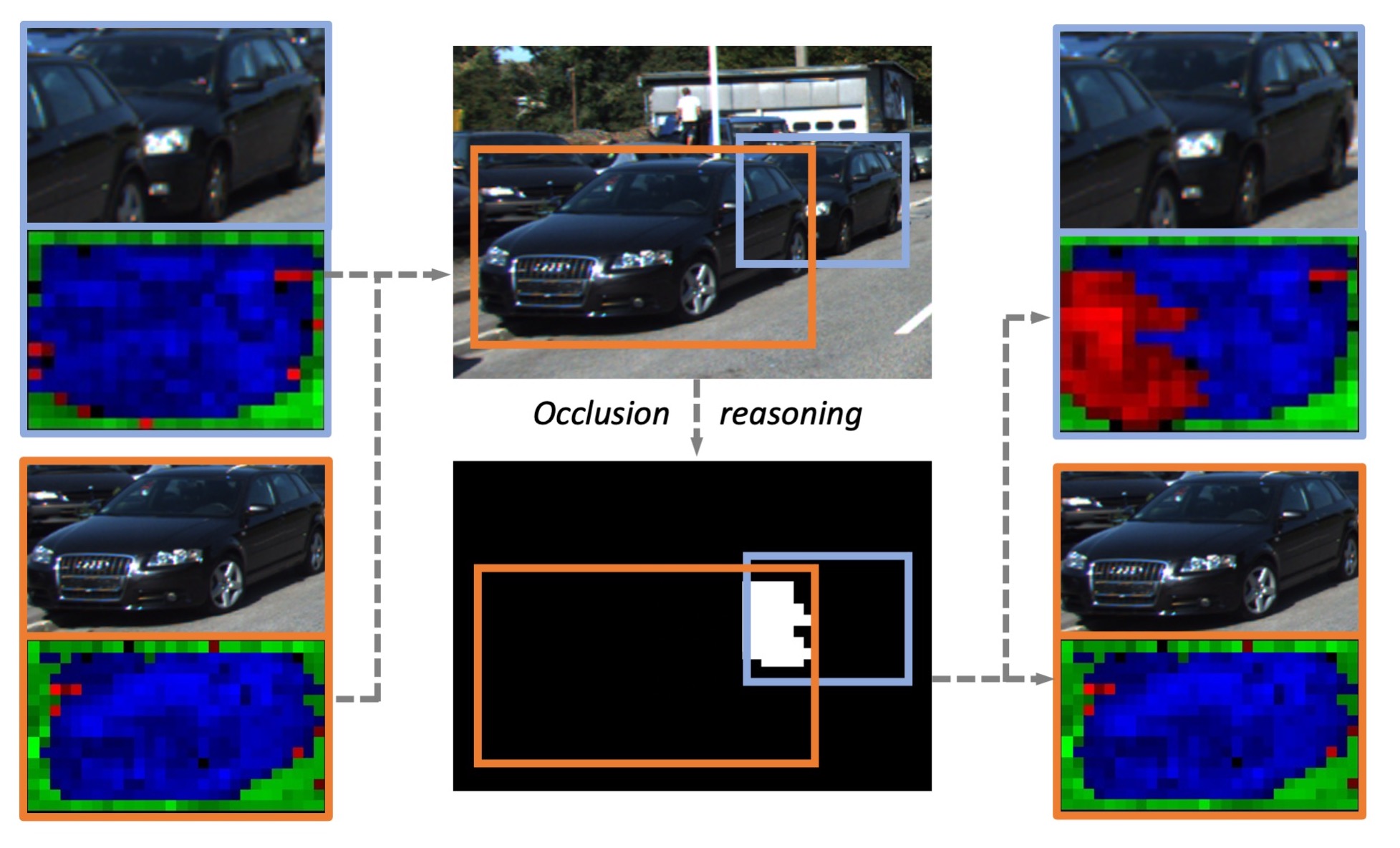}
    \caption{Our proposed model corrects erroneous instance segmentations through multi-object reasoning. Left: Two input images that are processed independently. The segmentation results identify visible object parts in blue, invisible parts in red, and context in green. Note how in the top image, the model cannot identify the occlusion. Center: By enforcing consistency between segmentations of nearby objects, our model can identify conflicting segmentations (white area). Right: Reasoning about the occlusion order resolves the erroneous predictions.}
    \label{fig:intro}
\end{figure}

\section{Introduction}
Scenes in images most often depict multiple objects that partially occlude each other.
Recent studies \cite{hongru,kortylewski2019compositional} showed that deep networks are less robust at recognizing partially occluded objects compared to Humans. 
The main difficulties are raised by the combinatorial variability of the object ordering and positioning, as well as the fact that scenes can contain known and unknown object classes. 

One approach to address the problem of occlusion in deep networks is data augmentation \cite{randErasing, devries2017improved, yun2019cutmix, chen2020vehdetec}. While this increases the robustness of deep networks, the classification performance on partially occluded objects still remains substantially worse compared to non-occluded objects.
Recent work introduced compositional deep networks (CompositionalNets) and showed that these are more robust to partial occlusion compared to data augmentation approaches~\cite{compnet_cls_cvpr20,kortylewski2020ijcv,compnet_det_cvpr20}.
CompositionalNets are deep neural network architectures in which the fully connected classification head is replaced with a differentiable compositional model. 
The structure of the compositional model enables CompositionalNets to decompose images into objects and context, as well as to further decompose objects into their individual parts.
The generative nature of the compositional model enables it to segment objects and occluders \cite{sun2020weaklysupervised} and to recognize objects based on their non-occluded parts. 
However, CompositionalNets, as well as other popular architectures, treat each object in an image independently and do not explicitly exploit the mutual relationship of nearby objects. 

In this paper, we introduce a deep network for multi-object instance segmentation that is robust to occlusion and can be trained from bounding box supervision only.
Our work builds on and significantly extends CompositionalNets.
Specifically, we extend the generative model in CompositionalNets to allow for instance segmentation of multiple mutually occluding objects in an image.  
This multi-object generative model is hard to optimize because of the mutual dependencies between objects.
To solve this optimization efficiently, we introduce an Occlusion Reasoning Module (ORM) that takes as input the independent predictions of each objects label, the instance segmentation and the occluder segmentation (Figure \ref{fig:intro}). 
We proceed to estimate possibly erroneous predictions through an occlusion voting mechanism. 
During occlusion voting, each object in the image votes for every pixel in its bounding box if the pixel is occupied by the object or if is occluded. 
Pixels which receive ambiguous votes from multiple objects indicate segmentation errors. 
To correct these we leverage the occlusion order of overlapping bounding boxes based on the classification scores. 
The corrected instance and occlusion segmentation masks are fed back into the CompositionalNet to mask out those features that induced segmentation errors, and to improve the prediction of the object class.

We perform extensive experiments on the KITTI INStance dataset (KINS). We further introduce a synthetic dataset that comprises artificially generated images of partially occluded objects, which are generated by superimposing segmented objects from the KITTI. The synthetic generation of partially occluded images enables us to evaluate custom types of occlusion challenges such as: pair-wise occlusion, multi-object occlusion and mixed occlusion containing both known and unknown object classes as occluders.
Our experimental results highlight that reasoning about multi-object occlusion significantly enhances the robustness of deep networks as it enables them to detect erroneous feed-forward predictions and self-correct through reasoning about multi-object occlusion.
In summary, our contributions in this work are: 
\begin{enumerate}[wide]
    \item We introduce a  deep  network for multi-object instance segmentation that is robust to occlusion and can be trained from bounding box super-vision only. Specifically, our network defines a generative model of multiple objects and achieves enhanced robustness through reasoning about multi-object occlusion.
    \vspace{-.1cm}
    \item We introduce an Occlusion Reasoning Module (ORM) that enables efficient inference in generative models with multiple objects. In particular, it detects erroneous feed-forward predictions and and corrects them through reasoning about the occlusion order of objects.
    \vspace{-.1cm}
    \item We achieve state-of-the-art performance at instance segmentation under occlusion on the KITTI INStance (KINS) dataset.
    \vspace{-.1cm}
    \item We introduce an occlusion challenge generated from real-world segmented objects with accurate annotations and propose a taxonomy of occlusion scenarios that pose a particular challenge for computer vision.
\end{enumerate}

\begin{figure*}[htbp]
\centering
    \includegraphics[width=17cm]{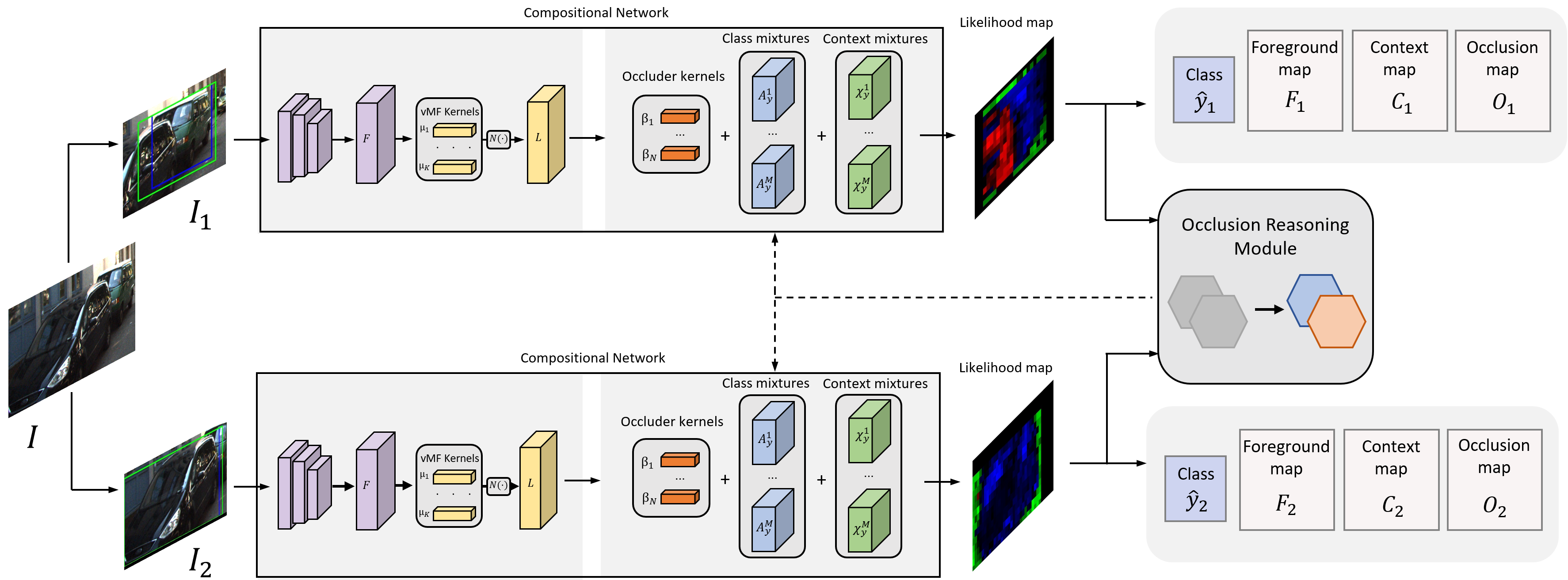}
    \caption{Proposed deep network architecture for multi-object instance segmentation under occlusion. Given an input image, we crop objects based on their bounding box. Each crop is processed by a Compositional Network to obtain independent estimates of the object class, instance segmentation and occlusion segmentation. Subsequently, these are processed by the multi-object reasoning module, which detects inconsistent segmentations and corrects them by taking into account the occlusion order of the objects. The corrected instance segmentation mask is used in a top-down manner to mask out occluded features, which, in turn, improves the classification score. Note we draw two Compositional Networks for illustrative purpose, in practice the images are processed sequentially by the same network.}
\label{pipeline}
\end{figure*} 

\section{Related Work}
\textbf{Occlusion reasoning.} A number of approaches have recently been proposed to integrate occlusion reasoning in areas including image classification \cite{kortylewski2020ijcv,xiao2020tdmpnet}, object detection \cite{compnet_det_cvpr20}, segmentation \cite{Gao2011,winn2006layout} and tracking \cite{Yang2014tracking}. 
Gao \textit{et al.}\cite{Gao2011} introduce binary variables to infer the visible cells in a bounding box. Hsiao and Hebert \cite{hsiao2014arbitrary_viewpoint} model occlusions by reasoning about 3D relationship of objects approximated by their bounding boxes. 
Recent works on pixel-level occlusion reasoning include
a probabilistic model proposed by George \textit{et al.}\cite{george2017captcha} that contains mutual occlusion inference on text-based CAPTCHAs by approximating MAP solution through message passing. Another probabilistic framework by Yang\textit{et al.}\cite{Yang2014tracking} introduce occlusion priori modeled by Markov random field to tackle mutual occlusion in object tracking task. Tighe \textit{et al.}\cite{tighe2014sceneparsing} introduce an inter-class occlusion prior to parse scenes and refine pixel-level labels. 
OFNet designed by Lu \textit{et al.}\cite{lu2019occ_relation} considers the relevance between occlusion contours and pixel orientations, but no semantic information is included. Zhan \textit{et al.}\cite{zhan2020de-occlusion} propose pair-wise order recovery by comparing the amodal mask completion of neighboring objects in a self-supervised way, while lack the ability of handling unknown occlusion. Our proposed architecture performs pixel-level occlusion reasoning and ensures the consistence of object shape by object-level occlusion order recovery. Note that we can handle both the unknown occlusion and multi-object occlusion at the same time.

\textbf{Weakly-supervised instance segmentation.} While instance segmentation performance was significantly advanced by CNN based architectures \cite{he2017mask, chen2019tensormask, Liu_2018_PANet, Chen_2019_cascade}, pixel-level semantic annotation is required for training by fully supervised methods. Weakly-supervised segmentation methods require only image-level supervision \cite{saleh2016built_in, oh2017exploiting} and bounding-box-level annotations \cite{rajchl2016deepcut, li2018panopic} to reduce the cost of dense labeling. DeepCut proposed by Rajchl \textit{et al.} \cite{rajchl2016deepcut} extends GrabCut \cite{rother2004grabcut} by training a CNN as classifier from bounding box annotations and address instance segmentation as energy minimisation problem based on conditional random fields. Zhou \textit{et al.}\cite{zhou2018classpeak} present an instance mask extraction by class response maps indicating visual cues with image-level supervision. Hsu \textit{et al.}\cite{hsu2019bbtp} address the problem as multiple instance learning task and estimate the foreground/background by generating positive/negative bags based on the sweeping lines of each bounding box. Amodal instance segmentation task were introduced more recently. 
Li \textit{et al.}\cite{li2016amodalseg} firstly presented a solution for amodal instance segmentation training with artificial occlusion. Other methods \cite{zhu2017semantic, qi2019kins, follmann2019endtoend} implement fully-supervised amodal mask completion. In this work, we build on the weakly-supervised instance segmentation  CompositionalNets\cite{sun2020weaklysupervised}, and generalize them to allow for reasoning about multi-object occlusion.

\begin{figure*}
\centering
    \includegraphics[width=17cm]{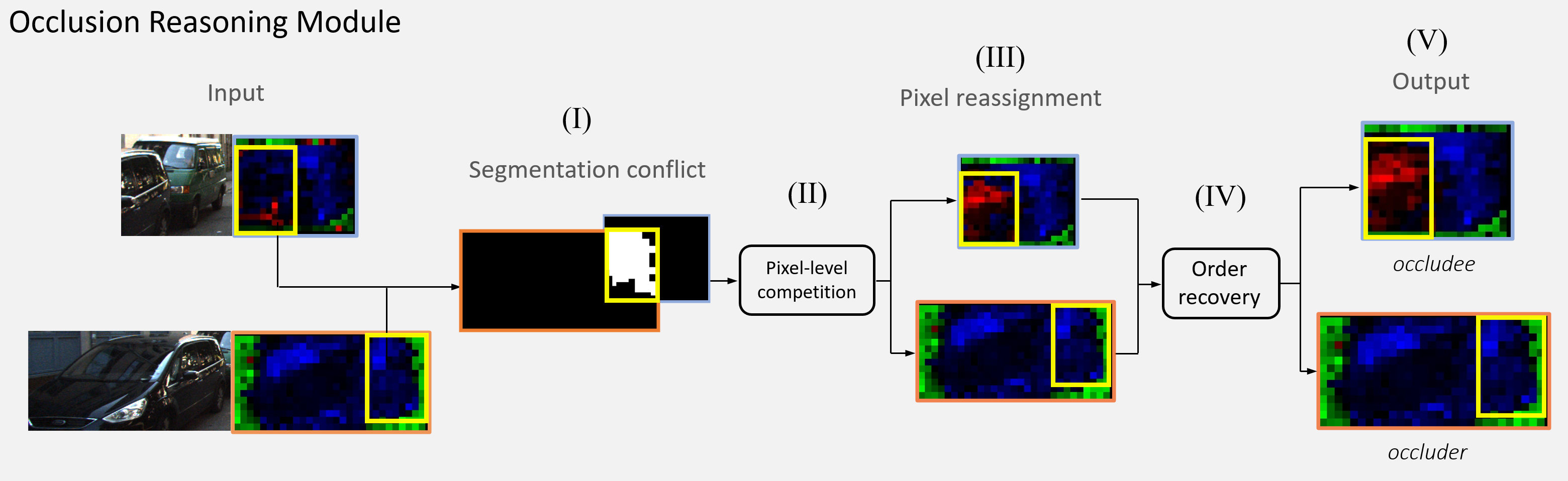}
    \vspace{-.3cm}
    \caption{Detailed method in Occlusion Reasoning Module (ORM). The input of ORM is the segmentation likelihood maps of a pair of neighboring objects. The map contains pixel-level prediction into foreground (blue pixels), background (green pixels) and occlusion (red pixels). Brighter pixel refers to higher likelihood on each position. Segmentation conflict is detected when one pixel is defined as foreground for both objects. Pixel-level competition is performed to solve the conflict and re-assign pixels. Pair-wise occlusion order recovery results from the competition, and the occludee's likelihood map is then updated accordingly.}
\label{pipeline_SCM}
\end{figure*}

\section{Robustness through Occlusion Reasoning}

\textbf{Notation.} The output of the layer $l$ in a DCNN is referred to as \textit{feature map} $F^l = \psi(I,\Omega) \in 
\mathbb{R}^{H \times W \times D}$, where $I$ and $\Omega$ are the input image and the parameters of the feature extractor, respectively. 
\textit{Feature vectors} are vectors in the feature map, $f_i^l \in \mathbb{R}^D$ at position $i$, where $i$ is defined on the 2D lattice of $F^l$ with $D$ being the number of channels in the layer $l$. 
We omit subscript $l$ in the following for clarity since the layer $l$ is fixed a priori in the experiments.
\subsection{Prior Work: CompNets for Single Objects}
\label{sec:prior}
CompositionalNets \cite{compnet_cls_cvpr20,kortylewski2020ijcv} are deep neural network architectures in which the fully connected classification head is replaced with a differentiable compositional model. 
In particular, the classification head defines a generative model $p(F|y)$ of the features $F$ for an object class $y$: 

\begin{equation}
\label{eq:vmf1}
    p(F|\Theta_y) {=} \sum_m \nu_m p(F|\theta^m_y), \hspace{.1cm}\nu_m \in\{0,1\}, \sum_{m=1}^M \nu_m {=} 1
\end{equation}

Here $M$ is the number of mixtures of compositional models per each object category and $\nu_m$ is a binary assignment variable that indicates which mixture component is active.
$\Theta_y{=} \{\theta^m_y {=} \{\mathcal{A}^m_y,\chi^m_{y},\Lambda\}|m{=}1,\dots,M\}$ are the overall compositional model parameters for the category $y$. The individual mixture components are defined as:
\begin{equation}
\label{eq:vmf2}
    p(F|\theta^m_y) = \prod_{i} p(f_i|\mathcal{A}_{i,y}^m,\chi^m_{i,y},\Lambda)
\end{equation}
Note how the distribution \textit{decomposes} the feature map $F$ into a set of individual feature vectors $f_i$. 
$\mathcal{A}^m_y=\{\mathcal{A}^m_{i,y}|i \in [H, W] \}$ and $\chi^m_y=\{\chi^m_{i,y}|i \in [H, W] \}$ are the parameters of the mixture components.

The feature likelihood is defined as composition of a foreground and a context likelihood:
\begin{align}
\label{eq:context_aware}
    	p(f_i|\mathcal{A}^m_{i,y},\chi^m_{i,y},\Lambda)= p(i|m,y) \hspace{0.1cm} p(f_i|\mathcal{A}^m_{i,y},\Lambda) \\
    	+ \hspace{0.1cm} (1-p(i|m,y)) \hspace{0.1cm} p(f_i|\chi^m_{i,y},\Lambda).
\end{align}{}
The parameters of the foreground and context likelihood are $\mathcal{A}^m_{i,y}$ and $\chi^m_{i,y}$ respectively.
$p(i|m,y)$ is a prior that models how likely a feature vector at position $i$ is to be located in the foreground. 
In particular, $\mathcal{A}^m_{i,y} = \{\alpha^m_{i,k,y}|k=1,\dots,K\}$ are mixture coefficients and $\Lambda = \{\lambda_k = \{\sigma_k,\mu_k \} | k=1,\dots,K \}$ are the parameters of von-Mises-Fisher distributions:
\begin{equation}
\label{eq:vmf3}
    p(f_i|\mathcal{A}_{i,y}^m,\Lambda) = \sum_k \alpha_{i,k,y}^m p(f_i|\lambda_k),
\end{equation}
\begin{equation}
\label{eq:vmfprob}
    p(f_i|\lambda_k) = \frac{e^{\sigma_k \mu_k^T f_i}}{Z(\sigma_k)}, ||f_i|| = 1, ||\mu_k|| = 1.
\end{equation}
The context likelihood  $p(f_i|\chi^m_{i,y},\Lambda)$ is defined accordingly.
Note that $K$ is the number of components in the vMF mixture distributions and  $\sum_{k=0}^K \alpha^m_{i,k,y} = 1$. $Z(\sigma_k)$ is the normalization constant. 
The priors $p(i|m,y)$ and likelihood parameters can be learned by segmenting the training images into foreground and background. We follow the approach introduced in \cite{compnet_det_cvpr20} which uses weakly supervised segmentation based on the bounding box annotation to segment the context from the object.
All model parameters $\{\Omega,\{\Theta_y\}\}$ can be trained end-to-end as discussed in \cite{compnet_cls_cvpr20, sun2020weaklysupervised}.

\textbf{Partial Occlusion.}
Compositional networks can be augmented with an outlier model to enhance their robustness to partial occlusion. 
The intuition is that at each position $i$ in the image either the object model $p(f_i|\mathcal{A}_{i,y}^m,\chi^m_{i,y},\Lambda)$ or an outlier model $p(f_i|\beta,\Lambda)$ is active:
\begin{align}
	&p(F|\theta^m_y,\beta)\hspace{-0.075cm} =\hspace{-0.075cm} \prod_{i} p(f_i|\beta,\Lambda)^{1-z^m_i} p(f_i|\mathcal{A}^m_{i,y},\Lambda)^{z^m_i}.\label{eq:occ}
\end{align}
The binary variables $\mathcal{Z}^m=\{z^m_i \in \{0,1\} | i \in \mathcal{P}\}$ indicate if the object is occluded at position $i$ for mixture component $m$. 

The outlier model is defined as: 
\begin{align}
p(f_i|\beta,\Lambda) = \sum_{k} \beta_{n,k} p(f_i|\sigma_k,\mu_k).
\end{align}
Note that the model parameters $\beta$ are independent of the position $i$ in the feature map and thus the model has no spatial structure. The parameters of the occluder models $\beta$ are learned from clustered features of random natural images that do not contain any object of interest \cite{compnet_cls_cvpr20}. 

\textbf{Instance segmentation with CompositionalNets.}
Sun et al.\cite{sun2020weaklysupervised} showed that instance segmentation can be achieved with CompositionalNets by simply comparing the likelihood terms of the model.
In particular, we can predict the pixel-wise labels to be foreground $\mathcal{F}$, context $\mathcal{C}$ or occlusion $\mathcal{O}$ by computing the respective likelihoods:
\begin{align}
    p(f_i = \mathcal{O}) &=p(i|m,y) \hspace{0.1cm} p(f_i|\beta,\Lambda)\\
    p(f_i = \mathcal{F},y) &=p(i|m,y) \hspace{0.1cm} p(f_i|\mathcal{A}^m_{i,y},\Lambda)\\
    p(f_i = \mathcal{C},y) &=(1 - p(i|m,y))  \hspace{0.1cm} p(f_i|\chi^m_{i,y},\Lambda)
\end{align}

\subsection{Compositional Networks for Multiple Objects}
The main limitation of Compositional Networks is that they assume only one object is present in an image.
They can be trivially generalized to multiple objects by treating each object independently \cite{compnet_cls_cvpr20, compnet_det_cvpr20}. 
However, assuming independence between objects neglects the relations between them and leads to inconsistencies in the segmentation results. For example Figure \ref{pipeline_SCM} shows how two objects with overlapping bounding boxes both predict that they are visible in the overlapped region. 
Whereas, it is clear that only one object can be visible per pixel in an image.

In this work, we aim to resolve such inconsistencies by enabling deep networks to reason about multi-object occlusion. In particular, we generalize the generative model in compositional networks to multiple objects by extending the model likelihood:
\begin{equation}\label{eq:multi}
    p(F|\theta^m_{y_1},\dots,\theta^m_{y_N},\beta) = \prod_i \prod_{n=1}^{N+1} p_n(f_i)^{z_{i,n}},
\end{equation}
with $\sum_n z_{i,n} {=} 1$ and $z_{i,n}{\in}\{0,1\}$. This generalized likelihood includes $n{=}\{1,\dots,N\}$ object models $p_n(f_i){=}p(F|\theta_{y_n}^m,\beta)$, which correspond to the number of objects in the image, and the outlier model $p_{N+1}(f_i){=}p(f_i|\beta,\Lambda)$. Note that, by the design of the likelihood, only one object model can be active at any location $i$ in the feature map $F$. Maximizing the model likelihood defined in Equation \ref{eq:multi} is difficult because it involves multiple objects and the visibility at each pixel $z_{i,n}$ depends on the visibility of the neighboring pixels. We solve this complex optimization problem by introducing a multi-object reasoning module into the architecture of CompositionalNets.

\subsection{Reasoning about Multi-Object Occlusion}
In this section, we introduce a deep  network for multi-object instance segmentation that is robust to occlusion and can be trained from bounding box super-vision only.
To make our discussion concise, we constrain ourselves in this section to images that contain two objects, where one partially occludes the other. 
Note, however, that our model trivially extends to multiple objects. 

\textbf{Feed-forward extraction of likelihood maps.} Our proposed network architecture is illustrated in Figure \ref{pipeline}. We draw two CompositionalNet architectures to enhance the clarity of the illustration, in practice they are sequentially processed by the same network. 
The objects in the input image $I$ are cropped based on their bounding box and first independently processed by a CompositionalNet. For each image crop $I_1, I_2$ we obtain a class prediction $\hat{y}_1,\hat{y}_2$ and three likelihood maps that encode the foreground, context and occlusion likelihood in the feature map:
\begin{align}
    \mathcal{F}_1= \{p(f_i = \mathcal{F},\hat{y}_1)|\forall i\in \mathcal{P}\}\\
    \mathcal{C}_1=\{p(f_i = \mathcal{C},\hat{y}_1)|\forall i\in \mathcal{P}\}\\
    \mathcal{O}_1=\{p(f_i = \mathcal{O})|\forall i\in \mathcal{P}\}
\end{align}
we compute $\mathcal{F}_2,\mathcal{C}_2,\mathcal{O}_2$ respectively.
We illustrate the likelihood maps throughout the paper as a two-dimensional heat map that is color coded. 
In particular, we visualize at each pixel which likelihood has the highest value by coloring occluder red, foreground blue and context green. The pixel intensity encodes the difference between the three likelihood terms. For dark pixels all likelihoods have similar values, hence indicating that the model is uncertain, whereas at bright pixels one likelihood is clearly higher compared to the other.
As shown in Figure \ref{pipeline_SCM}, instance segmentation based on the likelihood maps can be incorrect in the region where the two bounding boxes overlap (yellow box), particularly, for the occluded object. 
In practice, we observe that errors occur most often when an object is occluded by another object of the same category.
As discussed earlier, this is caused by the fact that the segmentation is performed independently of other objects in an image, and in addition also per pixel independently.
While these independence assumptions enable an efficient feed-forward inference, they neglect important relationships in images.
For example in the overlapping region of the bounding boxes in Figure \ref{pipeline_SCM}, we want all pixels to be assigned to the same object. It is very unnatural to treat every pixel independently.
We propose a multi-object reasoning module that resolves such segmentation conflicts by taking into account additional relationships between objects at minimal computational overhead. 

\textbf{Pixel-level competition.} Figure \ref{pipeline_SCM} illustrates the pipeline of the occlusion reasoning module. We first detect segmentation conflicts as those image pixels are classified as foreground by both object models (Figure \ref{pipeline_SCM}, I). We denote this conflict set as $\mathbb{C}$.
Note that for the occludee (the occluded object), some of the feature vectors in the occluded region are mis-classified as foreground, however, their likelihoods are lower compared to those of the occluder at the same pixel (indicated by the intensity of the color). 
\begin{figure}
\centering
\begin{subfigure}{0.23\textwidth}
\includegraphics[width=\textwidth]{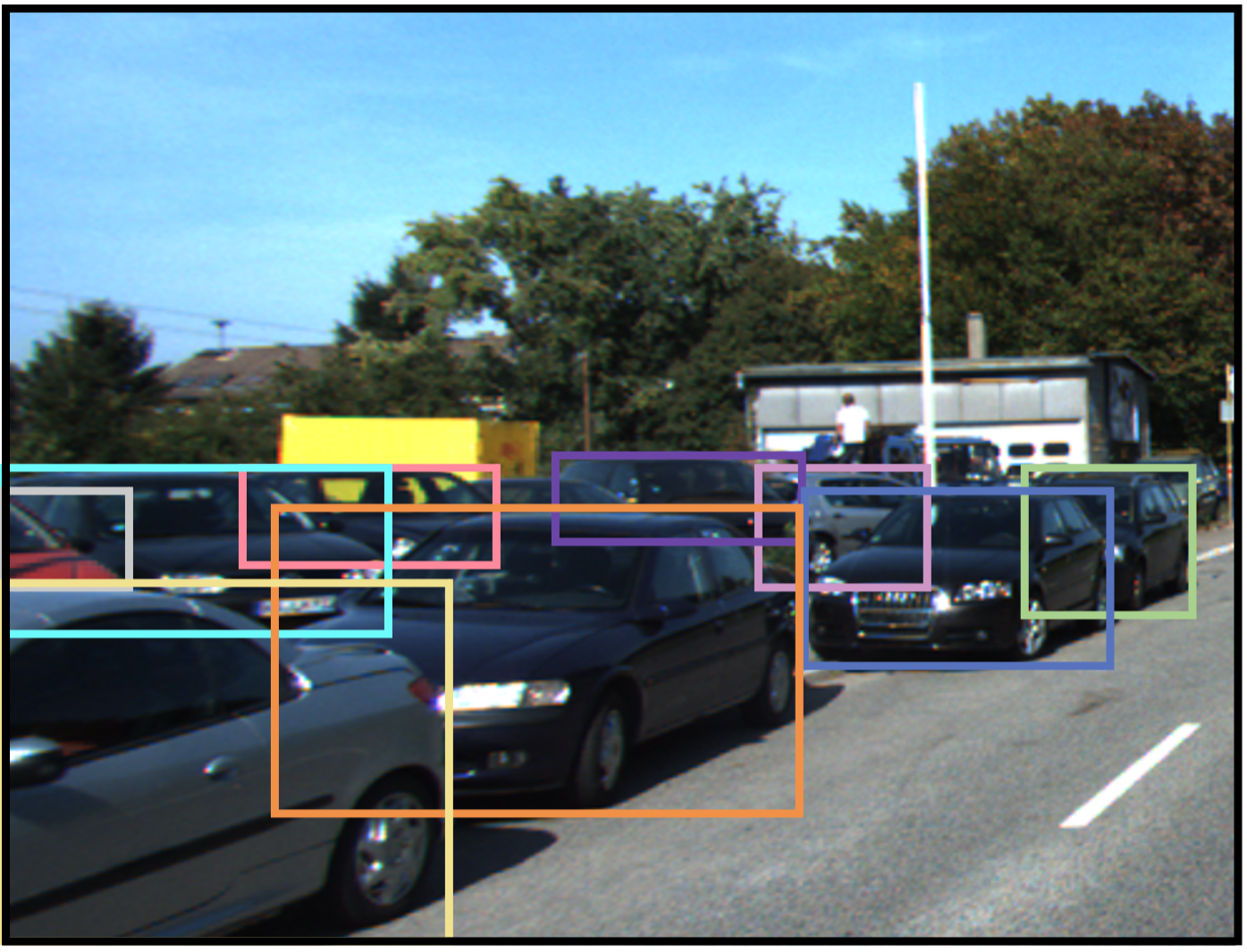}
\subcaption*{Input image}
\end{subfigure}
\hfill
\begin{subfigure}{0.23\textwidth}
\includegraphics[width=\textwidth]{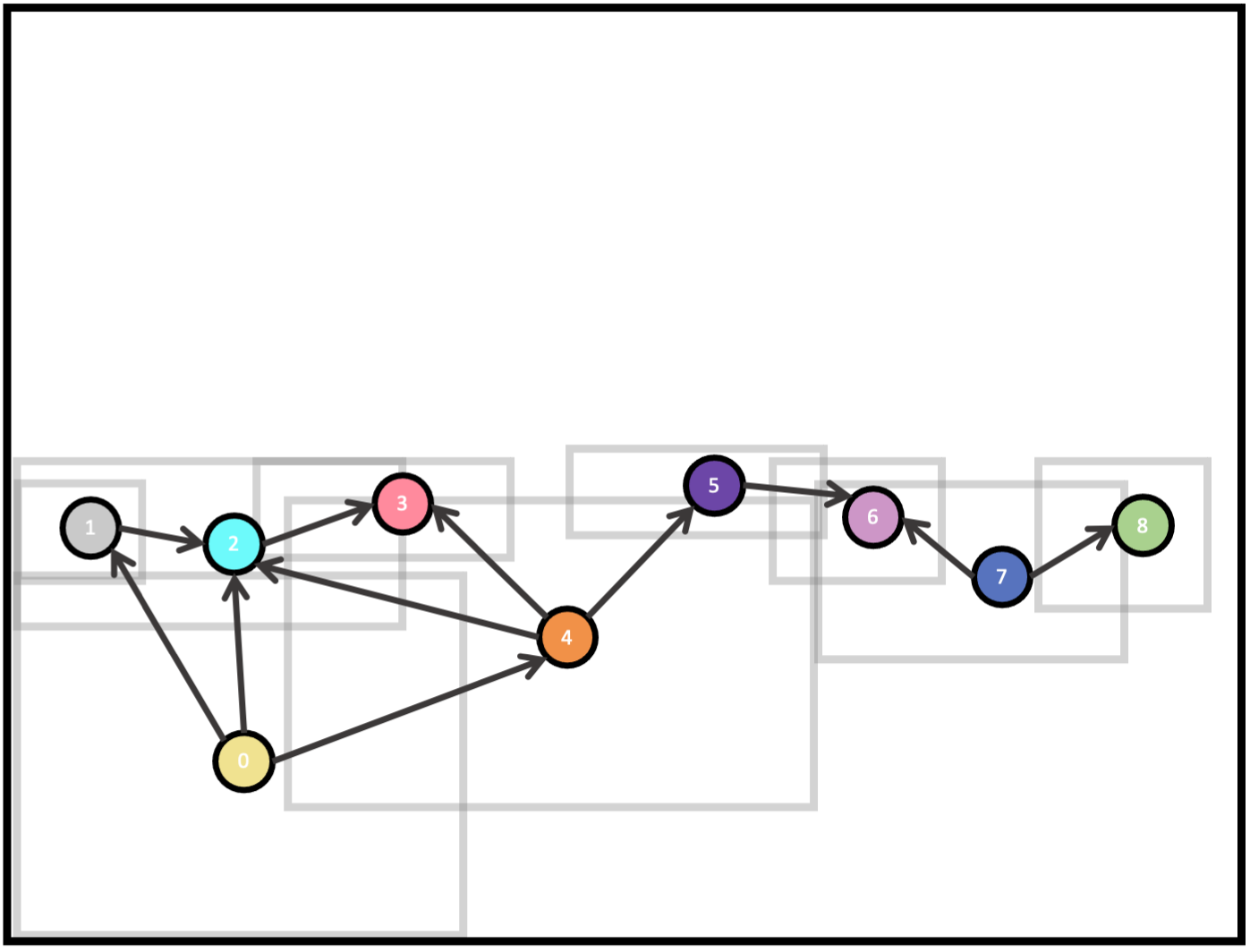}
\subcaption*{Order graph}
\end{subfigure}
\caption{Occlusion order graph recovered by the proposed network. Left: Input image with bounding boxes; Right: Occlusion order graph, where the direction of the arrows indicates occlusion. Note how the correct ordering is recovered in a very challenging occlusion scenario.}
\label{order_graph}
\end{figure} 

We exploit this by taking into account the relationship between both objects. In particular, we assign the feature vector $f_{i}$ to one of the two objects by comparing their foreground likelihoods (Figure \ref{pipeline_SCM}, II):
\begin{align}
    z_{i,1}{=} 
    \begin{cases}
        1,\hspace{-.1cm} &\text{if }p(f_i {=} \mathcal{F},\hat{y}_1) > \max\{p(f_i {=} \mathcal{F},\hat{y}_2),p(f_i {=} \mathcal{O})\} \\
        0,\hspace{-.1cm}  &\text{otherwise}
    \end{cases} 
\end{align}
We compute the visibility variables of the second object $z_{i,2}$ and the outlier model $z_{i,3}$ accordingly.
Using the estimated visibility variables, we can re-assign each pixel in the corresponding segmentation maps (Figure \ref{pipeline_SCM}, III).

\begin{figure}
    \centering
    \includegraphics[width=0.48\textwidth]{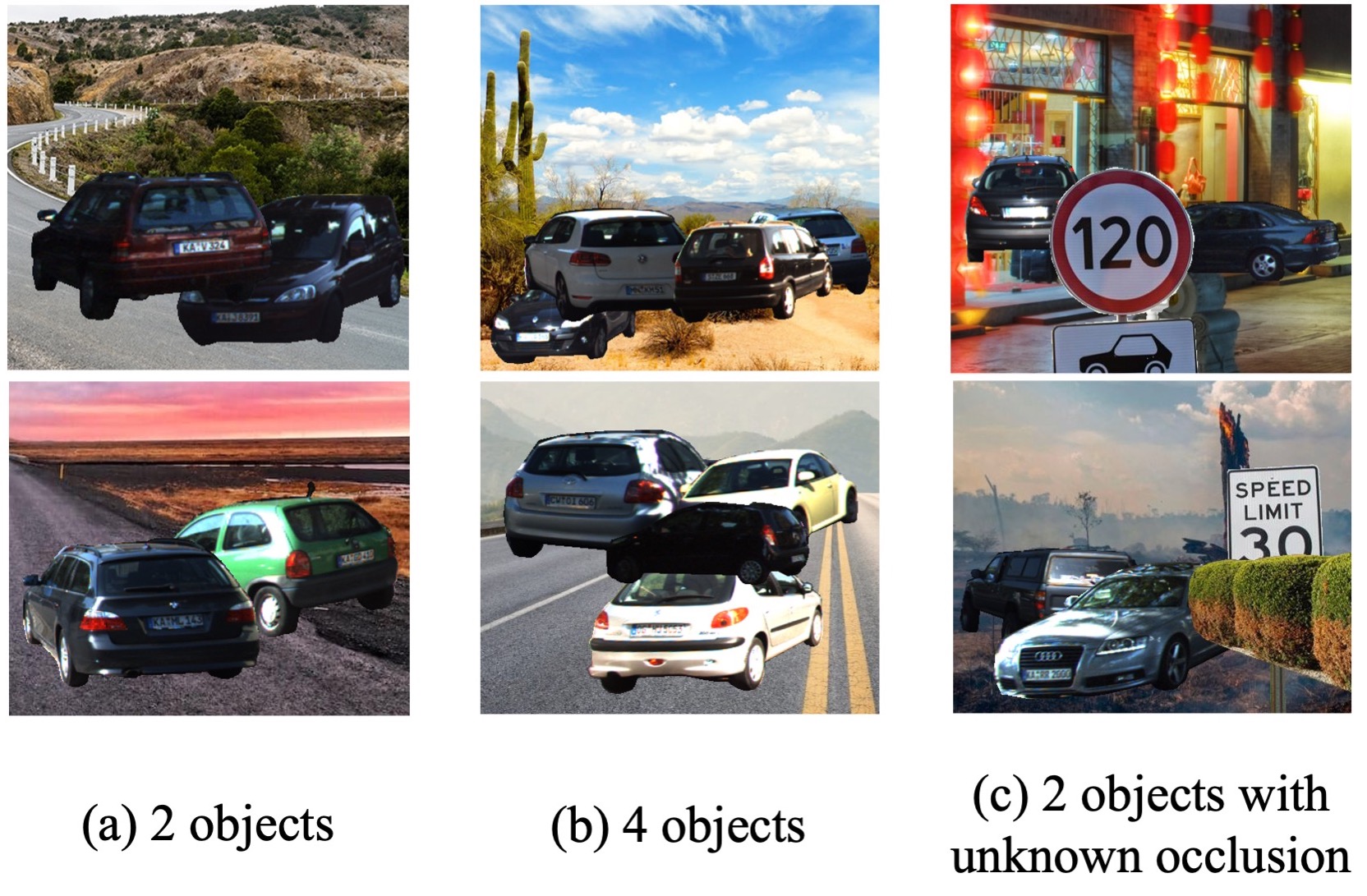}
    \caption{We synthesis three challenging occlusion scenarios: Occlusion with (a) 2 objects; (b) 4 objects with complex inter-object occlusion; and (c) multi-object occlusion including objects with of previously unseen occlusion.
    }
    \label{fig:synth_data}
\end{figure}

\textbf{Order recovery.} At this stage, each pixel has been updated independently. However, it is natural to assume that in the overlapping region of two bounding boxes only one object is the foreground object, whereas the other object is in the background. 
Hence all pixels should be assigned to either of the two objects. To encode this property we estimate the occlusion order between the objects (Figure \ref{pipeline_SCM}, IV).
Specifically, we estimate the occlusion order $R(I_1,I_2)$ by comparing the number of pixels assigned to each object in the region of the segmentation conflict:
\begin{equation}
\vspace{-.2cm}
    R(I_1,I_2)= 
    \begin{cases}
        1,  & \sum_{i\in \mathbb{C}} z_{i,1} > \sum_{i\in \mathbb{C}} z_{i,2}\\
        -1, & \text{otherwise .}
    \end{cases} 
\end{equation}
Figure \ref{order_graph} illustrates the effectiveness of this approach at recovering the occlusion order, even in challenging, multi-object occlusion scenarios.
Using the predicted occlusion order $R(I_1,I_2)$, we reassign the visibility variables $z_{i,n}$ in al "all or nothing" manner, such that all the variables that are not assigned to the outlier model are assigned to the object in the front.  
Figure \ref{pipeline_SCM} illustrates how the multi-object occlusion reasoning benefits the instance segmentation compared to the input segmentation which was achieved by processing the images independently.

\textbf{Self-correction through occlusion updates.} Compared to the occlusion variables $\mathcal{Z}^m$, which were estimated in the feed-forward stage, the newly estimated visibility variables $z_{i,n}$ take into account the knowledge of neighboring objects and their occlusion order graph.
This newly acquired knowledge can subsequently be used to recompute the model likelihood $p(F|\Theta_y)$ of the occluded object, by replacing the occlusion variables in Equation \ref{eq:occ}.
As our experimental results demonstrate, this top-down refinement enables CompositionalNets to correct miss-classifications that were induced by wrongly estimated occlusion variables. This will particularly improve the classification performance of occluded objects by a large margin. 
We repeat the self-correction through multi-object reasoning recurrently as the updated classification score can lead to changes in the assignment of the mixture models, and hence can lead to improved segmentations.

\begin{table}
\centering
\tabcolsep=0.08cm
\begin{tabular}{|l|c||c|c|c|c|c|c|c|}
\hline
 & Mask & L0 & L1 & L2 & L3 & Mean \\
\hline
Mask R-CNN   & \cmark & 85.8 &81.5 &72.7 &51.9 &73 \\
\hline
\hline
CompNet      & \xmark & 75.8 & 67.7 & 44.4 & 23.3 & 64.3 \\

Ours (iter=2) &  \xmark & \textbf{75.9} & \textbf{69.2} & \textbf{54.0} & \textbf{34.6} & \textbf{67.2} \\
\hline
\end{tabular}
\vspace{.2cm}\\
\begin{tabular}{|l|c|c|c|c|c|c|c|}
\hline
         & Mask & L0 & L1 & L2 & L3 & Mean \\
\hline
PCNet-M  & \cmark & 83.1 & 77.5 & 68.5 & 51.6 & 70.2 \\
\hline
\hline
BBTP     & \xmark & 77.9 & 71.6 & 67 & 67.8 & 71.1 \\
CompNet  & \xmark & 76.6 & 76.1 & 75.9 & 74.7 & 76.2 \\
Ours (iter=2) & \xmark & \textbf{76.9} & \textbf{76.4} & \textbf{76.5} & \textbf{76.5} & \textbf{76.7} \\
\hline
\end{tabular}
\caption{Modal and amodal instance segmentation on the KINS dataset (top and bottom). 
We compare to fully-supervised Mask R-CNN, self-supervised PCNet-M, weakly-supervised BBTP, and CompNets with and without ORM. Occlusion levels L1-L3 are defined as: L1: 1\%-30\%, L2: 30\%-60\%, L3: 60\%-90\% of the object is occluded.}
\label{table_kins}
\end{table}

\begin{figure*}
    \centering
    \includegraphics[width=0.98\textwidth]{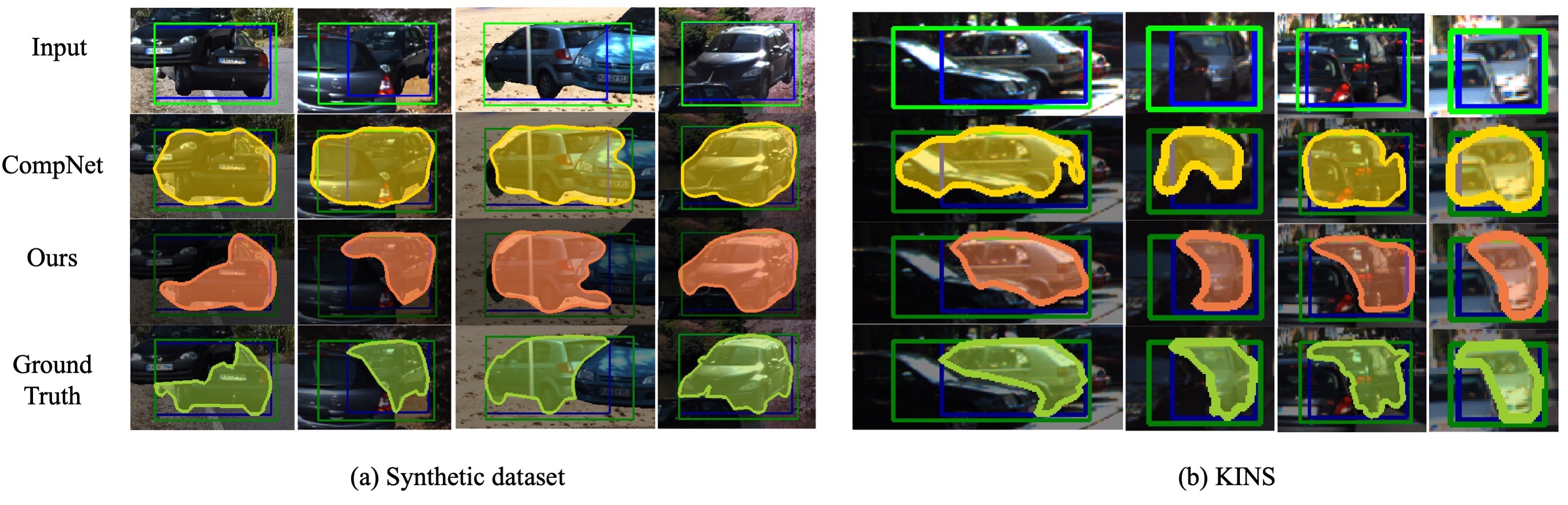}
    \vspace{-.3cm}
    \caption{Qualitative results for modal segmentation on KINS and images from our occlusion challenge. The top row show the input images including bounding box annotations. Images in the second row are generated by the baseline CompNet, and the third row shows the results by our CompNet with multi-object ORM. The last row shows the ground truth.}
    \label{fig:visual_seg_synth}
\end{figure*}

\section{Experiments}
We evaluate our deep network for multi-instance segmentation under occlusion on the KINS dataset and on an artificially generated occlusion challenge dataset. We will present experimental results of weakly-supervised modal and amodal instance segmentation, and ablate the occlusion reasoning module for order recovery.

\subsection{Datasets}
\textbf{KINS.} The KINS dataset \cite{qi2019kins} is augmented from KITTI \cite{geiger2013vision} with more instance pixel-level annotation for 8 categories including amodal instance segmentation and relative occlusion order. Amodal instance segmentation aims at segmenting the complete instance shape, even when the object is only partially visible. The dataset contains 7474 images for training and 7517 for testing. 

\textbf{Occlusion Challenge.} 
The amodal segmentation predicted on 2D real-world images by human judgements is still subjective and imprecise. Synthetic datasets are created to generate pixel-accurate annotations for the invisible parts of objects. Some generate 2D images from synthetic 3D scenes, e.g., DYCE \cite{Ehsani_seGan} provides natural configuration of objects in indoor scenes and SAIL-VOS\cite{Hu_SAILVOS} provides densely labeled video data extracted from the photo-realistic game GTA-V. These datasets contain natural object boundaries, while being deficient in photo-realistic textures. Others like \cite{li2016amodalseg} superimposing objects over other images to create artificial occlusion with real-world textures.
For the purpose of studying different types of occlusion challenges, we introduce a dataset with custom artificially generated occlusion scenarios. We crop non-occluded objects from images in KITTI based on their segmentation mask, and place them in images with random backgrounds (Figure \ref{fig:synth_data}). 

Since only complete visible objects are selected and the exact shape of each object is available, our occlusion challenge provides more accurate annotation for amodal masks compared with the human estimated masks in KINS. 
Most importantly, the synthetic nature of the dataset allows us to design challenging scenarios occlusion scenarios. In particular, we propose three types of occlusion challenges: 1) The basic and simplest occlusion scenario includes two objects, where one occludes the other. 2) A much more complex occlusion relationship is defined when four objects occlude each other with different amounts partial occlusion. Recovering the occlusion order and modal as well as amodal segmentation requires significant reasoning processes, even for humans. 3) Another challenging scenario is defined when the occluders contain a mixed set of object classes, some of which are known at training time, while are some are natural objects that are not part of the training data, such as street signs, bushes, and etc.

\begin{table*}
\centering
\tabcolsep=0.08cm
\setlength{\tabcolsep}{1.5mm}{
\begin{tabular}{|l|c c c c c|c c c c c|c c c c c|}
\hline
 &\multicolumn{5}{c|}{2 Objects} & \multicolumn{5}{c|}{4 Objects} &\multicolumn{5}{c|}{2 Objects + Unknown Occlusion} \\
\hline
Occ Level & L0 & L1 & L2 & L3 & Mean & L0 & L1 & L2 & L3 & Mean & L0 & L1 & L2 & L3 & Mean\\
\hline
Mask R-CNN    &88.2	&86.3	&69.1	&58.2	&82.3	&88.7	&88	&74.8	&63	&78.6	&90.5	&86.8	&72.2	&57.1	&76.7 \\
\hline
\hline
CompNet       & 77.8 & 67.3 & 51.0 & 26.3 & 66.9 &  \textbf{76.7} & 67.1 & 50.2 & 26.1 & 56.0 & \textbf{78.9} & 72.2 & 57.8 & 36.0 & 63.6 \\
Ours (iter=1) & \textbf{78.0} & \textbf{75.3} & 65.4 & 45.6 & 72.9 &  75.2 & \textbf{72.9} & 61.9 & 43.0 & 65.0 & 77.9 & \textbf{73.3} & \textbf{62.0} & \textbf{41.7} & \textbf{65.8} \\
Ours (iter=2) & \textbf{78.0} & \textbf{75.3} & \textbf{65.7} & \textbf{47.2} & \textbf{73.1} &  75.2 & \textbf{72.9} & \textbf{62.2} & \textbf{44.0} & \textbf{65.3} & 78.0 & \textbf{73.3} & \textbf{62.0} & \textbf{41.7} & \textbf{65.8} \\
\hline
\end{tabular}}
\vspace{.2cm}\\
\setlength{\tabcolsep}{1.5mm}{
\begin{tabular}{|l|c c c c c|c c c c c|c c c c c|}
\hline
 &\multicolumn{5}{c|}{2 Objects} & \multicolumn{5}{c|}{4 Objects} &\multicolumn{5}{c|}{2 Objects + Unknown Occlusion} \\
\hline
Occ Level & L0 & L1 & L2 & L3 & Mean & L0 & L1 & L2 & L3 & Mean & L0 & L1 & L2 & L3 & Mean\\
\hline
PCNet-M    &82.4	&81	&69.3	&47	&70	&87.2	&79.3	&63.7	&41.3	&67.9  &- &- &- &- &- \\
\hline
\hline
BBTP          & \textbf{80.5} &73.6 &69.5 &72.8 &74.1  &\textbf{80.5} &71.9 &64 &66 &70.6  &\textbf{83.7} &77.3 &67.9 &60.6 &72.4 \\
CompNet       & 78.0 & 76.6 & 75.0 & 72.1 & 76.7 & 77.3 & 75.4 & 74.1 & 71.4 & 74.8 & 78.4 & \textbf{78.1} & 76.1 & 71.9 & 76.5\\
Ours (iter=1) & 79.9 & \textbf{80.0} & 79.2 & 77.7 & \textbf{79.7} & 78.6 & 78.9 & 78.1 & 76.6 & 78.2 & 78.6 & 78.0 & \textbf{76.2} & \textbf{72.1} & \textbf{76.6} \\
Ours (iter=2) & 79.9 & \textbf{80.0} & \textbf{79.3} & \textbf{78.1} & \textbf{79.7} & 80.0 & \textbf{80.0} & \textbf{79.3} & \textbf{78.1} & \textbf{79.5} & 78.5 & \textbf{78.1} & \textbf{76.2} & \textbf{72.1} & \textbf{76.6} \\
\hline
\end{tabular}}
\caption{Modal and amodal instance segmentation on our occlusion challenge (top and bottom respectively). We compare to fully-supervised Mask R-CNN, PCNet-M, and weakly-supervised BBTP, and CompNets with and without ORM. Occlusion levels L0-L3 are defined as: L0: 0\%-1\%, L1: 1\%-30\%, L2: 30\%-60\%, L3: 60\%-90\% of the object are are occluded. Comparison between different times of occlusion reasoning iteration is also reported. Note that PCNet-M by design cannot handle unknown occlusion, and therefore cannot be applied in the last challenge.}
\label{table_synth}
\end{table*}

\begin{table}
\centering
\setlength{\tabcolsep}{1mm}{
\begin{tabular}{|l|c c| c c| c c|}
\hline
 &\multicolumn{2}{c|}{2 objects}  &\multicolumn{2}{c|}{4 objects}  &\multicolumn{2}{c|}{2 + unknown} \\
\hline
        & Modal & Amodal & Modal & Amodal & Modal & Amodal \\
 \hline
 NOD & 70.5 & 77.8 & 58.5 & 75.2 & 65.0 & 76.5 \\
 \hline
 OD & \textbf{73.1} & \textbf{79.7} & \textbf{65.3} & \textbf{79.5} & \textbf{65.8} & \textbf{76.6}  \\
 \hline
\end{tabular}}
\caption{Ablation study for order recovery. We compare the modal and amodal instance segmentation results for each occlusion challenge with and without order recovery. }
\label{table_abla_order_recovery}
\end{table}

\subsection{Implementation Details}
\textbf{Baselines.} We implement Mask R-CNN\cite{he2017mask} as baseline method for the modal instance segmentation. We further compare to CompositionalNets \cite{sun2020weaklysupervised} with and without our proposed multi-object reasoning module. We apply multi-object reasoning either with one reasoning iteration (iter=1) or recurrently with two iterations (iter=2).
Note that the CompositionalNets perform segmentation in a weakly-supervised manner from bounding box annotations only.
We compare our method with BBTP \cite{hsu2019bbtp} and PCNet-M \cite{zhan2020de-occlusion}. BBTP uses a bounding box tightness prior to perform weakly-supervised instance segmentation using box-level annotations. PCNet-M performs amodal mask completion in a self-supervised manner. PCNet-M is trained to recover the amodal mask with a given artificially occluded modal mask. In contrast, our model predicts amodal masks with bounding box supervision only and is capable of handling both known and unknown occluder classes. 

\textbf{Training setup.} We follow the training strategy as proposed in \cite{compnet_cls_cvpr20,sun2020weaklysupervised}. CompositionalNets are trained from the feature activations of a ResNeXt-50 \cite{he2016resnext} model that is pretrained on ImageNet\cite{deng2009imagenet} and fine-tuned on the respective datasets. 
We set the number of mixture components to $M=8$. We train for $60$ epochs using SGD with momentum $r=0.9$ and a learning rate of $lr=0.01$. 

\subsection{Instance segmentation under Occlusion}
\textbf{Modal segmentation.} We report modal instance segmentation performance in the top Tabulars in Table \ref{table_kins} on KINS and Table \ref{table_synth} on our occlusion challenge. Four occlusion levels of objects are defined as: L0: 0\%-1\%, L1: 1\%-30\%, L2: 30\%-60\%, L3:60\%-90\% of the object area being occluded. To prevent the performance from being affected by a poor bounding box prediction, all models are given the ground truth amodal bounding boxes during training and testing. 
For the KINS data, we observe that the fully supervised method outperforms weakly-supervised methods. However, our proposed multi-object extension with occlusion reasoning manages to significantly reduce the gap between weakly supervised methods and the fully supervised baseline. We outperform the baseline CompNet performance in every occlusion level, especially in higher occlusion levels by L2((9.6\%) and L3 (11.3\%) in terms of \textit{mIoU}. 
We observe similar performance patterns on the data for our occlusion challenge. 
While the CompNet performs similarly for the first and third occlusion challenge, its performance drops significantly when four objects mutually occlude each other compared to the other two scenarios. 
Our multi-object occlusion reasoning module enables CompNets to close this performance gap. Overall, the multi-object reasoning improves the segmentation results in all occlusion levels and for all challenge scenarios, and in particular for high occlusion levels L2 and L3. 

\textbf{Amodal segmentation.} We report amodal instance segmentation in the bottom Tabulars in Table \ref{table_kins} and Table \ref{table_synth}. Note that the self-supervised PCNet-M requires the modal mask as supervision to learn amodal mask completion. From the results, we observe that our model outperforms all other weakly-supervised methods in all levels of occlusion on the KINS data as well as in the occlusion challenge. We even surpass the mask-supervised PCNet-M in overall performance by 6.5\% in \textit{mIoU}.

In summary, with the ability of reasoning about multi-object occlusion, our proposed ORM significantly improves the robustness to occlusion compared with primary CompositionalNet. It achieves accurate instance segmentation in challenging occlusion scenarios (Figure \ref{fig:visual_seg_synth}. Our weakly-supervised model even outperforms mask-supervised methods in terms of amodal instance segmentation.

\subsection{Ablation study}
In Table \ref{table_abla_order_recovery}, we verify the effectiveness of the order recovery by evaluating modal and amodal segmentation results on our occlusion challenge. We perform experiments without pair-wise order (NOD), and with our predicted pair-wise order (OD). The results demonstrate the benefit of the order recovery, since per pixel competition cannot always correctly indicate the occluder and the occludee.

\section{Conclusion}
In this paper, we introduced a deep network for multi-object instance segmentation that is robust to occlusion and can be trained from bounding box supervision only.
In particular, our network defines a generative model of multiple objects and achieves enhanced robustness through reasoning about multi-object occlusion.
We further extended our architecture with an occlusion reasoning module that enables efficient inference  in  generative  models  with  multiple objects. 
In particular, it detects erroneous feed-forward predictions and and corrects them through reasoning about the occlusion order of objects.
Our experiments demonstrate the robustness of our proposed deep network for instance segmentation under occlusion on the KITTI INstance dataset and a dataset with synthetic occluders.

\textbf{Acknowledgements.} We gratefully acknowledge funding support from ONR N00014-18-1-2119, ONR N00014-20-1-2206 and the Swiss National Science Foundation (P2BSP2.181713). 

{\small
\bibliographystyle{ieee_fullname}
\bibliography{main}
}

\end{document}